\documentclass[11pt,a4paper]{article}
\usepackage[hyphens]{url}
\usepackage[hyperref]{eacl2021}
\usepackage{times}
\usepackage{latexsym}

\usepackage{times}
\usepackage{latexsym}

\usepackage[utf8]{inputenc}
\usepackage[english]{babel}
\usepackage{soul}
\usepackage{multirow}
\usepackage{booktabs}
\usepackage{tikz}
\usepackage[super]{nth}
\usepackage[ruled,vlined]{algorithm2e}

\usepackage{microtype}

\aclfinalcopy %

\newcommand\approach{sparsely factored~}
\newcommand\Approach{Sparsely Factored~}

\title{Sparsely Factored Neural Machine Translation}

\author{Noe Casas\textsuperscript{\textdagger}\textsuperscript{\textasteriskcentered},~~~
Jose A. R. Fonollosa\textsuperscript{\textasteriskcentered},~~~
Marta R. Costa-jussà\textsuperscript{\textasteriskcentered} \\
\textsuperscript{\textdagger} Lucy Software, United Language Group \\
\textsuperscript{\textasteriskcentered} TALP Research Center, Universitat Politècnica de Catalunya \\
\texttt{\{noe.casas,jose.fonollosa,marta.ruiz\}@upc.edu} \\
}

\date{}

\makeatletter
\expandafter\def\expandafter\UrlBreaks\expandafter{\UrlBreaks%
  \do\a\do\b\do\c\do\d\do\e\do\f\do\g\do\h\do\i\do\j%
  \do\k\do\l\do\m\do\n\do\o\do\p\do\q\do\r\do\s\do\t%
  \do\u\do\v\do\w\do\x\do\y\do\z\do\A\do\B\do\C\do\D%
  \do\E\do\F\do\G\do\H\do\I\do\J\do\K\do\L\do\M\do\N%
  \do\O\do\P\do\Q\do\R\do\S\do\T\do\U\do\V\do\W\do\X%
  \do\Y\do\Z}
\makeatother

\begin{document}
\maketitle
\begin{abstract}

The standard approach to incorporate linguistic information
to neural machine translation systems
consists in maintaining separate vocabularies for each of the
annotated features to be incorporated (e.g. POS tags,
dependency relation label), embed them, and then aggregate
them with each subword in the word they belong to.
This approach, however, cannot easily accommodate
annotation schemes that are not dense for every word.

We propose a method suited for such a case,
showing large improvements in out-of-domain data,
and comparable quality for the in-domain data. 
Experiments are performed in morphologically-rich languages
like Basque and German, for the case of low-resource scenarios.

\end{abstract}

\section{Introduction}

Domain shift is one of the main challenges yet to overcome
by neural machine translation (NMT) systems
\citep{koehn2017sixchallenges}. This problem happens when
using an MT system to translate data that is different
from the data used to train it, mainly regarding its domain
(e.g. the MT system was trained on news data but is then
used to translate biomedical data). The problem consists
in a drop in the translation quality with respect to
translations of in-domain text.

Injecting linguistic information has been used in the past to
improve the translation quality of NMT
systems. The improvements obtained for in-domain data are
normally small, while those obtained for out-of-domain text
are usually larger. The most frequent and straightforward approach
to inject linguistic information into NMT systems is to use
annotation systems to obtain word lemmas and part-of-speech (POS)
tags. These pieces of information are then attached as ``factors''
to each subword in the original word \cite{sennrich2016linguistic}.
In this scheme, however, it is assumed that each word has a value
for each of the possible factors. We refer to these as ``dense''
linguistic annotation schemes.

Nevertheless, not all linguistic annotations are dense. Some examples
of morphologically-rich language features that are not dense
include noun cases and verb conjugations, where only some type
of words can be tagged with such kind of information. These
``sparse'' linguistic annotation schemes cannot be easily accommodated
in factored NMT architectures.

In this work, we propose an approach to inject sparse linguistic
annotations into NMT systems. We refer to it as \approach NMT.

\section{Related Work} \label{sec:related}

Factored MT was first proposed by
\citet{koehn-hoang-2007-factored} for phrase-based statistical
machine translation systems (SMT) \citep{koehn-etal-2003-statistical}.
They propose to break down the translation process into
three stages: first translating input lemmas into output lemmas,
then mapping the source factors to target factors, and then
generating surface from the outcomes of the two previous stages,
using phrase-based SMT framework for each of the three stages.

The factored approach was first introduced in NMT by
\citet{sennrich2016linguistic}. In their approach,
lemmas, morphological features (case, number and gender
for nouns, person, number, tense and aspect for verbs),
POS tags and dependency labels are used as linguistic
information to enrich the source-side of an NMT system.
These pieces of word-level information
were attached to each of the subwords belonging to
the associated word, for the source-side sentences.
Apart from the linguistic information, subwords were
tagged with information about whether they are at the
beginning, at the middle or at the tail of the word.
\citet{hoang2016improving} also proposed the factored NMT
approach, and studied the effect of different attention
variants on it.
While the factored NMT approach was tested on the sequence-to-sequence
with attention architecture \citep{bahdanau2015seq2seq},
\citet{armengol2020enriching} studied its applicability
to the Transformer model \citep{vaswani2017attention}.

An analogous approach was used for the target side
by \citet{garcia2016factored} and \citet{burlot2017targetfactors}
to reduce the size of the word-level output vocabulary,
generating both the lemma and the morphological features that,
combined by means of an external morphological tool,
rendered the specific surface form. A larger study over
different translation directions with morphologically-rich
languages was performed by \citep{garcia2020sparsity}.

\section{\Approach NMT} \label{sec:approach}

\begin{figure}[ht]
\centering
\includegraphics[width=\linewidth]{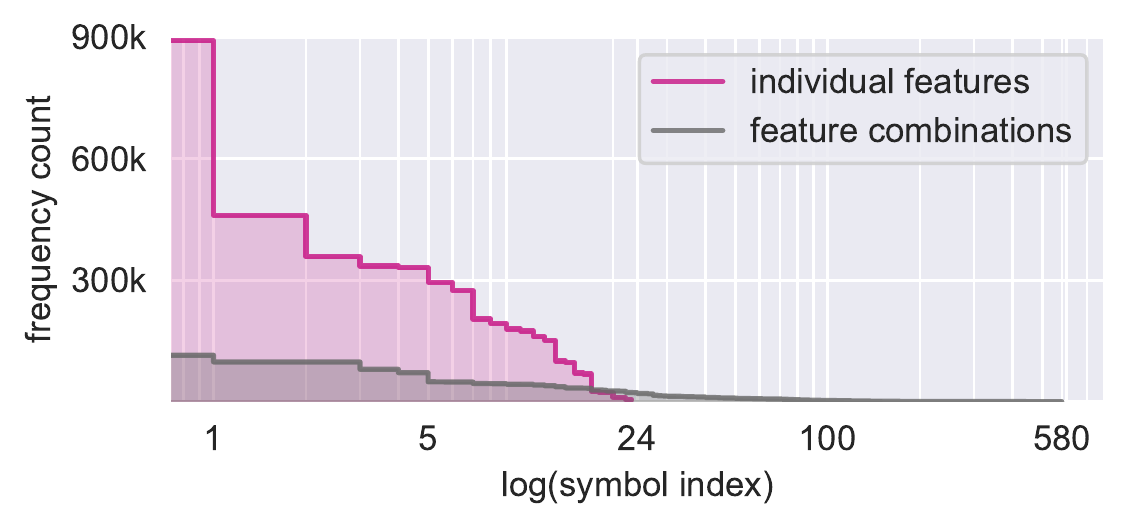}
\caption{Presence of the morphological information attributes
in the training data words. The grey histogram reflects the
frequency count of the different combinations of the morphological
features factors in the IWSLT14 de-en German data, while the
purple histogram reflects the frequency count of each
morphological feature value with the same textual data.
The morphological features were extracted with \texttt{ParZu}.
The X axis is expressed in logarithmic scale.
\label{fig:sparsity}}
\end{figure}

In our proprosed approach, instead of taking raw text as
input to translate, like normal NMT systems do,
we receive the text annotated by a linguistic
annotation system for the source-side. For training, in
the target side we take raw text. This aspect is the same
as in the work by \citet{sennrich2016linguistic}.
However, in their work, for sparse linguistic annotation schemes, like the
morphological features they use, each annotation is
a collection of attributes that may or may not be
present for each word type. The space of possible values
of the morphological features factor is large, as each word can have
a combination of such feature values, and a specific combination
may seldom appear in the training data, despite the fact
that each of its individual feature values may appear frequently.
This leads to a situation where many of the embedded vectors of the
morphological features factor are updated infrequently during
training. This fact is illustrated in Figure \ref{fig:sparsity},
where we show the frequency count of the morphological feature
combinations versus the frequency count of each individual
morphological feature, for the training split of one datasets used
in our experiments. In that figure, we can appreciate that
the number of different combinations is an order of magnitude
larger than the individual features (580 combinations vs. 24
individual feature values),
and that the frequency count is also multiple times lower.

Instead of taking each combination as a different factor value,
we propose to label each word based on the morphological feature
space instead of the morphological feature combination space.
For this, we keep an embedding table where each entry is a value
of a morphological feature. For instance, in the German sentence
``Wir brauchen Daten, keine Hilfe'', taken from the IWSLT14
validation data, in factored NMT pronoun ``Wir'' would be labeled
with the morphological feature combination
\texttt{1|Pl|\_|Nom} (first person, plural, nominative case),
while in \approach NMT the same word would be labeled with
three tags: \texttt{1}, \texttt{Pl} and \texttt{Nom}.

Also, regarding the use of word tokenization or subword tokenization
(e.g. Byte-Pair Encoding \citep[BPE;][]{sennrich2016bpe}), we
propose the following. Apart from the morphological feature
vocabulary described before, we also maintain a lemma-vocabulary;
when encoding text for \approach NMT, for each word we check if
it is a lemmatizable word (i.e. not a number, punctuation, etc) and if
its lemma is present in the lemma vocabulary. If it is, we
encode the word as the addition of the embedded vector of the lemma
plus the embedded vectors of every morphological feature the word had.
If the word could not be lemmatized or if the lemma is not present
in the lemma vocabulary, we tokenize the word using BPE, for which
we keep also an embedding table.
Therefore, our tokens can be either (lemma + morphological
features) or subwords. Once the text is encoded as a sequence
of embedded vectors, it is passed as input to a standard
Transformer model \citep{vaswani2017attention}.

Note that our proposal only affects the the embedding layer
of the encoder of an NMT architecture.
Therefore, it can be applied to both sequence-to-sequence with
attention or the Transformer.

We propose a further extension on top of the base variant
described before: we take a new hyperparameter, the ``linguistic dropout''
(LD), which represents the probability of using a subword tokenization
for a word instead of the (lemma + morphological features)
representation. During data preparation, both
the subword representation and the lemmatized representation (if available)
are prepared and, during training,
a sample of the Bernoulli distribution
with the LD probability determines which representation
is used for each word at batch creation time. The purpose
of LD is to make the model learn to handle the situation where there
is no linguistic information available (e.g. for out-of-vocabulary
words). Using LD, the subword token embeddings are
more frequently updated during training, leading to more robust
systems, especially on out-of-domain data.
This description is further enhanced with an algorithmic
presentation in Appendix \ref{sec:algorithm}.

\section{Experimental Setup}

In our experiments, we make use of morphologically-rich languages,
namely German and Basque, with low-resource scenarios, testing
with both in-domain and out-of-domain data.

German is a West Germanic Language with fusional morphology.
Its nouns are inflected in terms of number (singular and plural),
gender (masculine, feminine and neuter) and case (nominative, accusative,
genitive and dative).
Verbs inflect for person (\nth{1}, \nth{2} and \nth{3}),
number, mood (indicative, imperative, subjunctive, infinitive), voice,
tense (present, preterite, perfect, pluperfect, future, future perfect),
grammatical aspect, and completion status.
For the German experiments,
we use the IWSLT14 German$\rightarrow$English dataset
\citep{cettolo2014report} as training data. Its statistics are
shown in Table \ref{tab:iwslt14}. For the in-domain translation
quality evaluation, we used the mentioned dataset test split,
while for out-of-domain translation evaluation we used the
WMT17 biomedical test sets, namely the
English-German HimL test set%
\footnote{\url{http://www.himl.eu/test-sets}}. The preprocessing
used was the one recommended by fairseq for the IWSLT14 de-en
data\footnote{\url{https://github.com/pytorch/fairseq/tree/master/examples/translation}}, namely corpus cleaning, tokenization and lowercasing
with Moses scripts \citep{koehn2007moses}, and the BPE subword
vocabulary had 10k merge operations.

\begin{table}[ht]
\centering
\fontsize{10pt}{12pt}\selectfont
\setlength{\tabcolsep}{3pt}
\centerline{
\begin{tabular}{ c c c c c c }
\textbf{Corpus} & \textbf{Sents.} & \textbf{Words} & \textbf{Vocab.} & \textbf{Max.len.} & \textbf{Avg.len.} \\
\hline
German &   & 3.1M & 113k & 172 & 19.4\\
English & \raisebox{1.5 ex}[0pt]{160k} & 3.3M & 53k & 175 & 20.4\\
\end{tabular}}
\caption{IWSLT14 German-English training data stats.}
\label{tab:iwslt14}
\end{table}

Basque is a language isolate (not related to other languages),
with agglutinative morphology.
Its nouns take suffixes to express number (singular, plural, ``mugagabe'')
and case (nominative, ergative, genitive, local genitive, dative, allative, inessive, partitive, etc).
Verbs' surface forms differ based on the person of the subject,
direct object and indirect object (\nth{1}, \nth{2} and \nth{3}),
number (singular and plural), tense (present, past, future),
aspect (progressive and perfect)
and mood (indicative, subjunctive, conditional, potential and imperative).
For the Basque experiments, we use the EiTB news corpus
\citep{etchegoyhen2016exploiting}. Its statistics are shown
in Table \ref{tab:eitb}. We split the original data%
\footnote{\url{https://aholab.ehu.eus/metashare/repository/browse/basque-spanish-eitb-corpus-of-aligned-comparable-sentences/5f5bd836b6f111e6b004f01faff11afa8b95c93ec1214a338167e5074ee90d09/}} into
training, validation and test subsets. The test split was
used for in-domain translation quality evaluation, while
a sample of 1000 sentences of the Open Data Euskadi
IWSLT18 corpus \citep{jan2018iwslt} containing documents
from the Public Administration, was used for the 
out-of-domain translation evaluation. The preprocessing for
the data consisted in truecasing and tokenization (this preprocessing
was already applied in the original data),
and the BPE subword vocabulary had 15k merge operations.

\begin{table}[ht]
\centering
\fontsize{10pt}{12pt}\selectfont
\setlength{\tabcolsep}{3pt}
\centerline{
\begin{tabular}{ c c c c c c }
\textbf{Corpus} & \textbf{Sents.} & \textbf{Words} & \textbf{Vocab.} & \textbf{Max.len.} & \textbf{Avg.len.} \\
\hline
Basque &   & 10.1M & 345k & 318 & 18.3\\
Spanish & \raisebox{1.5 ex}[0pt]{550k} & 15.6M & 225k & 317 & 28.3\\
\end{tabular}}
\caption{EiTB Basque-Spanish training data statistics.}
\label{tab:eitb}
\end{table}

\begin{table*}[ht]
\centering
\fontsize{11pt}{15pt}\selectfont
\begin{tabular}{r c c  c c}
& \multicolumn{2}{c}{\textsc{\textbf{eu$\rightarrow$es}}} & \multicolumn{2}{c}{\textsc{\textbf{de$\rightarrow$en}}} \\
\textsc{Model}~ &  \textsc{in domain} & \textsc{out of domain} & \textsc{in domain} & \textsc{out of domain} \\
\hline
Without linguistic info~ & 29.8 & 19.9 & 34.8 & 3.2 \\
Factored~~ & 24.2 & 13.8 & 32.0 & 8.4 \\
\Approach~~ & 28.6 & 19.7 & 32.6 & 8.0\\
\Approach + LD~~ & 29.4 & 20.7 & 34.3 & 9.2\\
\end{tabular}
\caption{Translation quality (case-insensitive BLEU scores) of the proposed model (\Approach NMT,
with and without linguistic dropout) and baseline models: BPE without
linguistic information and Factored NMT.}
\label{tab:bleu}
\end{table*}

The linguistic information used for the experiments was obtained
with Lucy LT \citep{alonso2003comprendium}, a rule-based
machine translation (RBMT) system of transfer type. We took
the analysis of the source sentences generated in the intermediate
stages of the translation, which annotates each word with a bag
of linguistic language-specific features, covering all the
morphological and grammatical traits of the word.

We train \approach NMT systems with and without linguistic
dropout. For LD, we used $p=0.25$, that is, there is a 75\% probability of
using the (lemma + morphological features) representation, if available,
and 25\% probability of using the word's subword tokens instead.
The neural architecture used for out experiments was the
Transformer model.

We included two baseline systems as reference.
First, a vanilla Transformer model
with BPE vocabulary without any linguistic information.
Second, a factored NMT system
\citep{sennrich2016linguistic}.
The hyperparameter configurations used for the baselines and
the evaluated models can be found in Appendix \ref{sec:hyperparameters}.

For the German linguistic
information we used the \texttt{ParZu} annotation tool
\citep{sennrich2009parzu,sennrich2013parzu} (which was
the tool used by \citet{sennrich2016linguistic}), while for Basque
we use the analysis by Lucy LT.

In our experiments, we studied the translation quality in terms
of BLEU scores \citep{papineni2002bleu}, obtained with Moses'
\texttt{multi-bleu.perl} script after tokenizing with the
Moses tokenizer. Given that our datasets had been truecased/lowercased,
we compute the lower-case variant of the BLEU score.

\section{Results}

Table \ref{tab:bleu} shows the BLEU scores obtained by our \approach
NMT, with and without linguistic dropout, as well as the
baseline systems, for the \mbox{German (\textsc{de}) $\rightarrow$ English (\textsc{en})}
and \mbox{Basque (\textsc{eu}) $\rightarrow$ Spanish (\textsc{es})}
translation directions, both with in-domain and out-of-domain tests.

We can see that the factored NMT system in general performs worse than
the Transformer baseline without linguistic information. This can be
associated with the sparsity problem described in Section \ref{sec:approach}
and illustrated in Figure \ref{fig:sparsity}, which is especially relevant
for an agglutinative language like Basque, where the difference for in-domain
data is 5.6 BLEU points.

We can also appreciate that with \approach NMT without LD, we also suffer a loss
in translation quality with respect to the vanilla Transformer.
However, using \approach NMT, we have comparable translation quality
with respect to the vanilla Transformer for in-domain data, but for out-of-domain
data we improve 0.8 BLEU points for Basque and 6 BLEU points for German.

From these results, we understand that, without LD, the subword token embeddings
are under-trained. This problem is mitigated by the introduction of LD. The
results also suggest that the improvements can be larger in very low resource
scenarios, like the German experiments, with 160k sentences in the training data,
a much smaller size than Basque, with 550k. For in-domain
data, our approach suffers a small loss, 0.4-0.5 BLEU points, which is normally
considered comparable.

\section{Conclusion}

We proposed \approach NMT, which is an approach to inject linguistic information
in the source-side of NMT architectures, especially appropriate for
annotation schemes where the morphological tags are not applicable
to all word types, leading to sparseness of the training signal
in classical approaches like factored NMT. We also proposed linguistic
dropout, a complement to \approach NMT that improves the training signal
for the subword embeddings.

Our experiments showed that this approach maintains the
baseline translation quality, only with a minor loss, and
improves drastically the translation quality of out-of-domain
text when the system has been trained in a low-resource setting.

Our code is available as open source at \mbox{\url{https://github.com/noe/sparsely_factored_nmt}}.
\ifaclfinal

\section*{Acknowledgments}

This work is partially supported
by Lucy Software / United Language Group (ULG)
and the Catalan Agency for Management of University and Research
Grants (AGAUR) through an Industrial Ph.D. Grant.
This work also is supported in part by the Spanish
Ministerio de Economía y Competitividad, the European Regional
Development Fund through the postdoctoral senior grant
Ramón y Cajal and by the Agencia Estatal de Investigación
through the projects EUR2019-103819 and PCIN-2017-079. 

\fi

\bibliography{eacl2021}
\bibliographystyle{acl_natbib}

\clearpage

\appendix

\section*{Supplementary Material: Appendix}

\section{Hyperparameters} \label{sec:hyperparameters}

All the models presented in Table \ref{tab:bleu} use the
Transformer architecture. For the German$\rightarrow$English
models (trained on the IWSLT14 de-en dataset), we used the
hyperparameter configuration recommended by fairseq for that dataset%
\footnote{\url{https://github.com/pytorch/fairseq/tree/master/examples/translation}},
which is presented in Table \ref{tab:hyperparams1}. 

\begin{table}[ht]
\centering
\fontsize{11pt}{15pt}\selectfont
\begin{tabular}{r c}
\hline
num. layers & 6\\
num. heads  & 4\\
embed. size & 512\\
feedforward size & 1024\\
total batch size & 4096 \\
\hline
\end{tabular}
\caption{\label{tab:hyperparams1}
Hyperparameters German$\rightarrow$English.}
\end{table}

For the
Basque$\rightarrow$Spanish models (trained on the EiTB eu-es dataset),
we use the base Transformer
configuration, which is presented in Table \ref{tab:hyperparams2},
but with a smaller total batch size of 4096 to avoid overfitting.

\begin{table}[ht]
\centering
\fontsize{11pt}{15pt}\selectfont
\begin{tabular}{r c}
\hline
num. layers & 6\\
num. heads  & 8\\
embed. size & 512\\
feedforward size & 2048\\
total batch size & 4096 \\
\hline
\end{tabular}
\caption{\label{tab:hyperparams2}
Hyperparameters for Basque$\rightarrow$Spanish.}
\end{table}

The value used for LD used for our \approach NMT models
was $p=0.25$ (i.e. 75\% probability of
using the lemma + morphological features representation, if available,
and 25\% probability of using the word's subword tokens instead).

The hyperparameter tuning was done manually, trying less than 8
configurations, focusing on linguistic dropout and
number of attention heads. All experiments were performed on
a server with 4 nvidia 1080Ti GPUs.

All models were trained once, with no retraining.

\section{Implementation Details} \label{sec:algorithm}

For the vanilla Transformer we used
fairseq \citep{ott2019fairseq}.
For the factored NMT, we used OpenNMT \citep{klein2017opennmt} (which
supports token features) with its implementation of the Transformer.
For the \approach NMT system, we used a modified version of fairseq's 
Transformer.

\section{Detailed Encoding Approach}

In Section \ref{sec:approach} we presented the description of
the approach used to encode the sparsely factored annotation
schemes. The same information is presented in a more specific
way as Algorithm \ref{alg:encoding}, where we can see
that, for each word in a sentence, if the word is lemmatizable
and the linguistic dropout mechanism (LD) (i.e. avoiding encoding the
linguistic information with probability $p$) allows, the
word is encoded as a vector obtained by adding the embedded
vector of the word's lemma together with the embedded vector
of each of its factors, or otherwise the word is decomposed
into subwords with BPE andeach subword is embedded and concatenated
with the rest of the sentence token vectors.

\begin{algorithm}[ht]
\fontsize{9pt}{12pt}\selectfont
\SetAlgoLined
\KwResult{token\_vectors}
 token\_vectors = []\;
 \ForEach{word in sentence\_words}{
  \eIf{is\_lemmatizable(word) \& not LD}{
    word\_vector = $[0, 0, …, 0]$\;
    lemma = get\_lemma(word)\;
    lemma\_vector = embedding[lemma]\;
    word\_vector += lemma\_vector\;
    factors = get\_factors(word)\;
    \ForEach{factor in factors}{
      factor\_vector = embedding[factor]\;
      word\_vector += factor\_vector\;
    }
    token\_vectors.append(word\_vector)\;
   }{
    \ForEach{subword in BPE(word)}{
      subword\_vector~=embedding[subword]\;
      token\_vectors.append(subword\_vector)\;
    }
  }
 }
 \caption{\label{alg:encoding}Sparsely factored encoding}
\end{algorithm}

Note that the presented conceptual algorithmic approach is
adapted at the implementation level with appropriate
parallelization and adapted to work in the automatic
differentiation framework offered by \texttt{fairseq}.

\end{document}